\documentclass[a4paper,10pt]{article}

\usepackage[utf8]{inputenc}
\usepackage[a4paper,left=2.5cm,right=2.5cm,top=2.5cm,bottom=2.5cm]{geometry}
\usepackage[backend=biber, style=apa, natbib]{biblatex}
\usepackage[hidelinks]{hyperref}
\usepackage{graphicx}
\usepackage{bbm}
\usepackage{amsmath}
\usepackage{amsfonts}
\usepackage{amssymb}
\usepackage{array}
\usepackage[font=small,labelfont=bf]{caption}
\usepackage{comment}
\usepackage{xspace}
\usepackage{mathtools}
\usepackage[ruled,vlined]{algorithm2e}
\usepackage{authblk}
\usepackage{pythonhighlight}
\usepackage{empheq}
\usepackage{color}

\definecolor{boxedcolor}{rgb}{.88, .95, .95}

\newlength\mytemplen
\newsavebox\mytempbox
\makeatletter
\newcommand\nicebox{%
    \@ifnextchar[
       {\@nicebox}%
       {\@nicebox[0pt]}}
\def\@nicebox[#1]{%
    \@ifnextchar[
       {\@@nicebox[#1]}%
       {\@@nicebox[#1][0pt]}}
\def\@@nicebox[#1][#2]#3{
    \sbox\mytempbox{#3}%
    \mytemplen\ht\mytempbox
    \advance\mytemplen #1\relax
    \ht\mytempbox\mytemplen
    \mytemplen\dp\mytempbox
    \advance\mytemplen #2\relax
    \dp\mytempbox\mytemplen
    \colorbox{boxedcolor}{\hspace{1em}\usebox{\mytempbox}\hspace{1em}}}
\makeatother

\DeclarePairedDelimiter\ceil{\lceil}{\rceil}
\DeclarePairedDelimiter\floor{\lfloor}{\rfloor}
\DeclarePairedDelimiter\abs{\lvert}{\rvert}
\DeclareMathOperator{\sign}{sign}

\bibliography{references}

\newcommand{\loihi}{\texttt{Loihi}\xspace}
\newcommand{\nxsdk}{\texttt{NxSDK}\xspace}
\newcommand{\brian}{\texttt{Brian}\xspace}

\title{\texttt{Brian2Loihi}: An emulator for the neuromorphic chip Loihi using the spiking neural network simulator Brian}

\author[1,2,*]{Carlo Michaelis}
\author[1,2,*]{Andrew B. Lehr}
\author[1,2,*]{Winfried Oed}
\author[1,2]{Christian Tetzlaff}

\affil[1]{Department of Computational Neuroscience, University of G{\"o}ttingen, Germany}
\affil[2]{Bernstein Center for Computational Neuroscience, University of G{\"o}ttingen, Germany}
\affil[*]{These authors contributed equally.}

\date{}

\begin{document}

\maketitle
\vspace{-0.6cm}
\begin{abstract}
\noindent Developing intelligent neuromorphic solutions remains a challenging endeavour.
It requires a solid conceptual understanding of the hardware's fundamental building blocks.
Beyond this, accessible and user-friendly prototyping is crucial to speed up the design pipeline.
We developed an open source Loihi emulator based on the neural network simulator Brian that can easily be incorporated into existing simulation workflows.
We demonstrate errorless Loihi emulation in software for a single neuron and for a recurrently connected spiking neural network.
On-chip learning is also reviewed and implemented, with reasonable discrepancy due to stochastic rounding.
This work provides a coherent presentation of Loihi's computational unit and introduces a new, easy-to-use Loihi prototyping package with the aim to help streamline conceptualisation and deployment of new algorithms.
\end{abstract}



\section{Introduction}

Neuromorphic computing offers exciting new computational structures.
Decentralised units inspired by neurons are implemented in hardware \citep[reviewed by][]{schuman2017survey, young2019review, rajendran2019low}.
These can be connected up to one another, stimulated with inputs, and the resulting activity patterns can be read out from the chip as output.
A variety of algorithms and applications have been developed in recent years, including robotic control \citep{dewolf2016spiking, michaelis2020robust, stagsted2020towards, dewolf2020nengo}, spiking variants of deep learning algorithms, attractor networks, nearest-neighbor or graph search algorithms \citep[reviewed by][]{davies2021advancing}.
Moreover, neuromorphic hardware may provide a suitable substrate for performing large scale simulations of the brain \citep{thakur2018large, furber2016large}.
Neuromorphic chips specialised for particular computational tasks can either be provided as a neuromorphic computing cluster or be integrated into existing systems, akin to graphics processing units (GPU) in modern computers \citep{furber2014spinnaker, davies2021advancing}.
With the right ideas, networks of spiking units implemented in neuromorphic hardware can provide the basis for powerful and efficient computation.
Nevertheless, the development of new algorithms for spiking neural networks, applicable to neuromorphic hardware, is a challenge \citep{gruning2014spiking, pfeiffer2018deep, bouvier2019spiking}.

At this point, without much background knowledge of neuromorphic hardware, one can get started programming using the various software development kits available (e.g., \citealp{lin2018programming, michaelis2020pelenet, rueckauer2021nxtf, rhodes2018spynnaker, muller2020operating, muller2020extending, spilger2020hxtorch, sawada2016truenorth, bruderle2011comprehensive}).
Emulators for neuromorphic hardware \citep{luo2018fpga, valancius2020fpga, petrovici2014characterization, furber2014spinnaker} running on a standard computer or field programmable gate arrays (FPGA), make it possible to develop neuromorphic network architectures without even needing access to a neuromorphic chip (see e.g. NengoLoihi\footnote{\url{https://www.nengo.ai/nengo-loihi/}} and Dynap-SE\footnote{\url{https://code.ini.uzh.ch/yigit/NICE-workshop-2021}}).
This can speed up prototyping as the initialisation of networks, i.e.~distributing neurons and synapses, as well as the readout of the system's state variables on neuromorphic chips takes some time.
At the same time emulators transparently contain the main functionalities of the hardware in code and therefore provide insights into how it works.
With this understanding, algorithms can be intelligently designed and complex network structures implemented.

In the following, we introduce an emulator for the digital neuromorphic chip \loihi \citep{davies2018loihi} based on the widely used spiking neural network simulator \brian \citep{stimberg2019brian}.
We first dissect an individual computational unit from \loihi.
The basic building block is a spiking unit inspired by a current based leaky integrate and fire (LIF) neuron model \citep[see][]{gerstner2014neuronal}.
Connections between these units can be plastic, enabling the implementation of diverse on-chip learning rules.
Analysing the computational unit allows us to create an exact emulation of the \loihi hardware on the computer.
We extend this to a spiking neural network model and demonstrate that both \loihi and \brian implementations match perfectly.
This exact match means one can do prototyping directly on the computer using \brian only, which adds another emulator in addition to the existing simulation backend in the Nengo Loihi library.
This increases both availability and simplicity of algorithm design for \loihi, especially for those who are already used to working with \brian.
In particular for the computational neuroscience community, this facilitates the translation of neuroscientific models to neuromorphic hardware.
Finally, we review and implement synaptic plasticity and show that while individual weights show small deviations due to stochastic rounding, the statistics of a learning rule are preserved.
Our aim is to facilitate the development of neuromorphic algorithms by delivering an open source emulator package that can easily be incorporated into existing workflows.
In the process we provide a solid understanding of what the hardware computes, laying the appropriate foundation to design precise algorithms from the ground up.


\section{Loihi's computational unit and its implementation}

Developing a \loihi emulator requires precise understanding of how \loihi works.
And to understand how something works, it is useful to ``take it apart and put it back together again''.
While we will not physically take the \loihi chip apart, we can inspect the components of its computational units with ``pen and paper''.
Then, by implementing each component on a computer we will test that, when put back together, the parts act like we expect them to.
In the following we highlight how spiking units on \loihi approximate a variant of the well-known LIF model using first order Euler numerical integration with integer precision.
This understanding enables us to emulate \loihi's spiking units on the computer in a way that is straightforward to use and easy to understand.
For a better intuition of how the various parameters on \loihi interact, we refer readers to our neuron design tool\footnote{anonymized} for \loihi.
Readers familiar with \citet{davies2018loihi} and numerical implementations of LIF neurons may prefer to skip to Section~\ref{sec:computational_unit:summary}.

\subsection{\loihi's neuron model: a recap}

The basic computational unit on \loihi is inspired by a spiking neuron \citep{davies2018loihi}.
\loihi uses a variant of the leaky integrate and fire neuron model \citep{gerstner2014neuronal} (see Appendix~\ref{sec:appendix:lif-variant}).
Each unit $i$ of \loihi implements the dynamics of the voltage $v_i$
\begin{equation}\label{eq:loihi:voltage:basis}
    \frac{dv_i}{dt} = - \frac{1}{\tau_v} v_i(t)  + I_i(t) - v_i^{th} \sigma_i(t),
\end{equation}
where the first term controls the voltage decay, the second term is the input to the unit, and the third term resets the voltage to zero after a spike by subtracting the threshold. A spike is generated if $v_i > v_i^{th}$ and transmitted to other units to which unit $i$ is connected. In particular, $v$ models the voltage across the membrane of a neuron, $\tau_v$ is the time constant for the voltage decay, $I$ is an input variable, $v^{th}$ is the threshold voltage to spike, and $\sigma(t)$ is the so-called spike train which is meant to indicate whether the unit spiked at time $t$.
For each unit $i$, $\sigma_i(t)$ can be written as a sum of Dirac delta distributions
\begin{equation}\label{eq:spike-train}
    \sigma_i(t) = \sum_k \delta(t - t_{i,k}),
\end{equation}
where $t_{i,k}$ denotes the time of the $k$-th spike of unit $i$.
Note that $\sigma_i$ is not a function, but instead defines a \textit{distribution} (i.e.~\textit{generalised function}), and is only meaningful under an integral sign. 
It is to be understood as the linear functional $\langle \sigma_i, f \rangle := \int \sigma_i(t) f(t) \,dt = \sum_k f(t_{i,k})$ for arbitrary, everywhere-defined function $f$ (see \hyperlink{corollary:dirac}{Corollary 1} in Appendix~\ref{subsec:appendix:synaptic_response}).

Input to a unit can come from user defined external stimulation or from other units implemented on chip.
\citet{davies2018loihi} describe the behavior of the input $I(t)$ with
\begin{equation}\label{eq:current:loihi:paper:original}
    I_i(t) = \sum_{j} J_{ij} (\alpha_I * \sigma_j)(t) + I_i^\text{bias},
\end{equation}
where $J_{ij}$ is the weight from unit $j$ to $i$, $I_i^\text{bias}$ is a constant bias input, and the spike train $\sigma_j$ of unit $j$ is convolved with the synaptic filter impulse response $\alpha_I$, given by
\begin{equation}\label{eq:alphau}
    \alpha_I(t) = \exp\left(-\frac{t}{\tau_I}\right)\,H(t),
\end{equation}
where $\tau_I$ is the time constant of the synaptic response and $H(t)$ the unit step function.
Note that $\alpha_I(t)$ is defined differently here than in \citet{davies2018loihi} (see Appendix~\ref{subsubsec:appendix:synaptic_filter} for details).
The convolution from Equation~\ref{eq:current:loihi:paper:original} is a notational convenience for defining the synaptic input induced by an incoming spike train, simply summing over the time-shifted synaptic response functions, namely $(\sigma_i*f)(t) = \langle \sigma_i, \tau_t \tilde f \rangle = \sum_k f(t - t_{i,k})$, where $\tau_t f(x) = f(x-t)$ and $\tilde f(x) = f(-x)$ (see Appendix~\ref{subsec:appendix:synaptic_response}).

\subsection{Implementing \loihi's spiking unit in software}
\label{seq:loihi:comp:neuro}

From the theoretical model on which \loihi is based, we can derive the set of operations each unit implements with a few simple steps.
Using a first order approximation for the differential equations gives the update equations for the voltage and synaptic input described in the \loihi documentation.
Combined with a few other details regarding \loihi's integer precision and the order of operations, we will have all we need to implement a \loihi spiking unit in software.

\subsubsection*{Synaptic input}

From Equation~\ref{eq:current:loihi:paper:original} we see that the synaptic input can be written as a sum of exponentially decaying functions with amplitude $J_{ij}$ beginning at the time of each spike $t_{j,k}$ (see Appendix~\ref{subsec:appendix:synaptic_response}). 
In particular we have
\begin{equation}\label{eq:loihi:input:result}
I_i(t) = \sum_j J_{ij}\sum_k \exp\left(\frac{t_{j,k} - t}{\tau_I}\right)H(t-t_{j,k}) + I_i^\text{bias}.
\end{equation}

To understand the behavior of the synaptic input it is helpful to consider the effect of one spike arriving at a single synapse.
Simplifying Equation~\ref{eq:loihi:input:result} to just one neuron that receives just one input spike at time $t_1 = 0$, for $t \ge 0$ we get
\begin{equation}\label{eq:loihi:input:simplified}
    I(t) = J \cdot \exp\left(-\frac{t}{\tau_I}\right)
\end{equation}

and for $t < 0$, $I(t) = 0$. Each spike induces a step increase in the current which decays exponentially with time constant $\tau_I$.
Taking the derivative of both sides with respect to $t$ gives
\begin{align}\label{eq:loihi:input:diff}
    \frac{dI}{dt} &= -\frac{1}{\tau_I} \cdot I(t), \\
    I(0) &= J.
\end{align}
Applying the forward Euler method to the differential equation for $\Delta t = 1$ and $t \ge 0$, $t \in \mathbb{N}$ we get
\begin{equation} \label{eq:loihi:input:euler}
    I[t] = I[t-1] - \frac{1}{\tau_I} \cdot I[t-1] + J \cdot s[t],
\end{equation}
where $s[t]$ is zero unless there is an incoming spike on the synapse, in which case it is one.
Here, $s[0] = 1$ and $s[t] = 0$ for $t > 0$.
With this we have simply incorporated the initial condition into the update equation.
Note that we have switched from a continuous (e.g.~$I(t)$) to discrete (e.g.~$I[t]$) time formulation, where $\Delta t = 1$ and $t$ is unitless.

\loihi has a decay value $\delta^I$, which is inversely proportional to $\tau_I$, namely $\delta^I = 2^{12} / \tau_I$.
Swapping $\tau_I$ by $\delta^I$ reveals
\begin{equation}\label{eq:loihi:input:euler:delta}
    I[t] = I[t-1] \cdot (2^{12} - \delta^I) \cdot 2^{-12} + J \cdot s[t].
\end{equation}
The weight $J$ is defined via the mantissa $\tilde w_{ij}$ and exponent $\Theta$ (see Section~\ref{subsec:synaptic_weights}) such that the equation describing the synaptic input becomes (with indices)
\begin{empheq}[box={\nicebox[5pt]}]{equation}
    I_i[t] = I_i[t-1] \cdot (2^{12} - \delta^I) \cdot 2^{-12} + 2^{6+\Theta} \cdot \sum_j \left( \tilde w_{ij} \cdot s_j[t] \right),
\label{eq:loihi:current:original}
\end{empheq}
where $s_j[t] \in \{0,1\}$ is the spike state of the $j^{th}$ input neuron. Please note that Equation~\ref{eq:loihi:current:original} is identical to the \loihi documentation.

From this we can conclude that the implementation of synaptic input on Loihi is equivalent to evolving the LIF synaptic input differential equation with the forward Euler numerical integration method (see Figure~\ref{fig:simulations-match}A1).

\subsubsection*{Voltage}

It is straightforward to perform the same analysis as above for the voltage equation.
We consider the subthreshold voltage dynamics for a single neuron and can therefore ignore the reset term $v_i^{th} \sigma_i(t)$ from Equation~\ref{eq:loihi:voltage:basis}, leaving us with
\begin{equation}\label{eq:loihi:voltage:simple}
    \frac{dv}{dt} = - \frac{1}{\tau_v} v(t)  + I(t).
\end{equation}

Applying forward Euler gives
\begin{equation}\label{eq:loihi:voltage:simple:euler}
    v[t] = v[t-1] - \frac{v[t-1]}{\tau_v} + I[t].
\end{equation}

Again, to compare with the \loihi documentation we need to swap the time constant $\tau_v$ by a voltage decay parameter, $\delta^v$, which is inversely proportional to the time constant, the same as above for synaptic input.
Plugging in $\tau_v = 2^{12} / \delta^v$ leads to
\begin{equation}\label{eq:loihi:voltage:simple:euler:delta}
    v[t] = v[t-1] \cdot (2^{12} - \delta^u) \cdot 2^{-12} + I[t].
\end{equation}
By introducing a bias term, the voltage update becomes
\begin{empheq}[box={\nicebox[5pt]}]{equation}
    v_i[t] = v_i[t-1] \cdot (2^{12} - \delta^u) \cdot 2^{-12} + I_i[t] + I_i^\text{bias}.
\label{eq:loihi:voltage:original}
\end{empheq}

Equation~\ref{eq:loihi:voltage:original} agrees with the \loihi documentation.
Like the synaptic input, the voltage implementation on \loihi is equivalent to updating the LIF voltage differential equation using forward Euler numerical integration (see Figure~\ref{fig:simulations-match}A2).

\subsubsection*{Integer precision}
\label{sec:loihi:round_away}

\loihi uses integer precision.
So the mathematical operations in the update equations above are to be understood in terms of integer arithmetic.
In particular, for the synaptic input and voltage equations the emulator uses \textit{round away from zero}, which can be defined as
\begin{equation}
    x_\text{round} \coloneqq \sign(x) \cdot \ceil{ \abs{ x } }.
\end{equation}
where $\ceil{\cdot}$ is the ceiling function and $\sign(\cdot)$ the sign function.

\subsection{Summary} 
\label{sec:computational_unit:summary}

\begin{figure*}[t]
    \centering
    \includegraphics[width=0.99\textwidth]{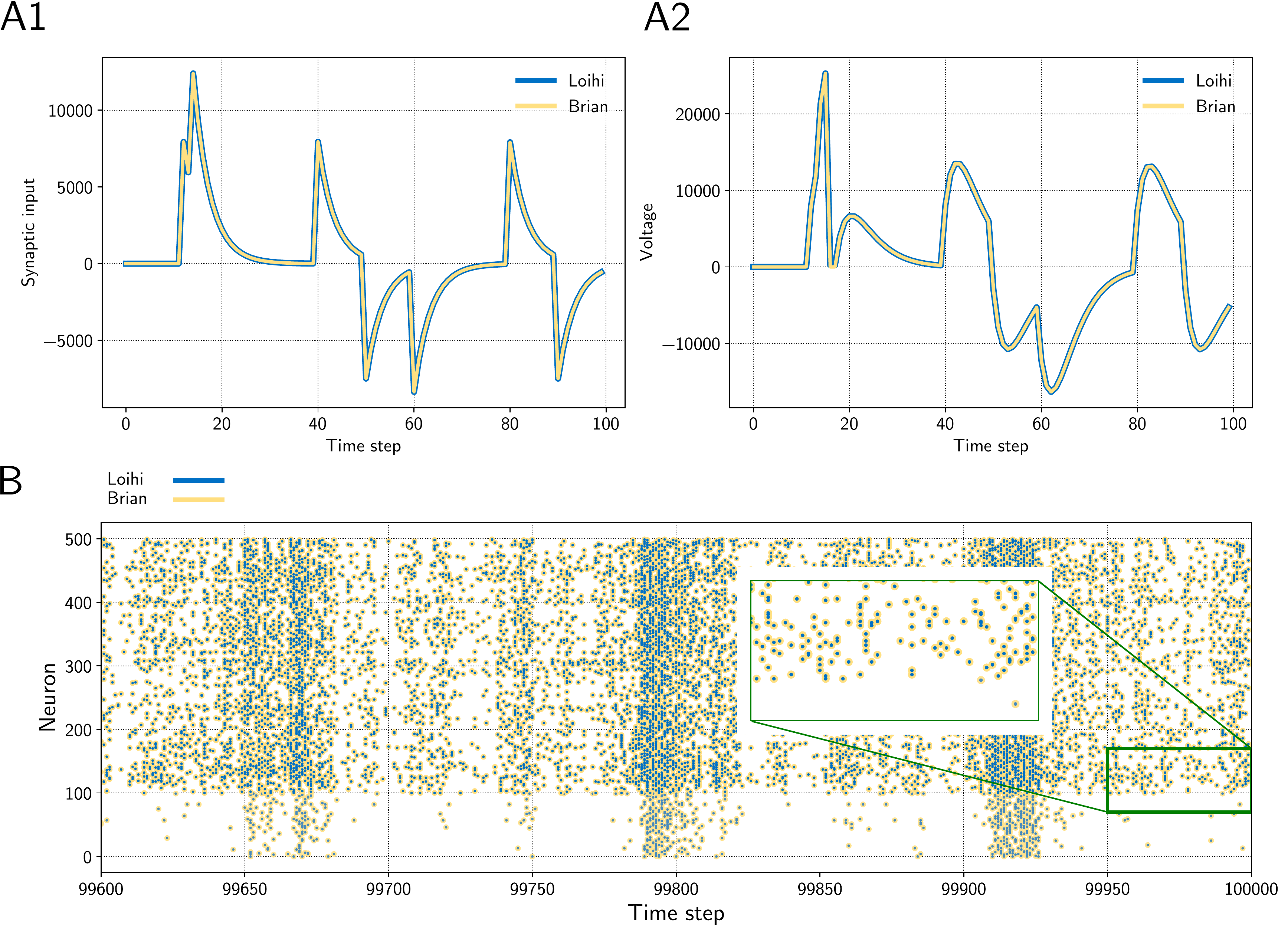}
    \caption{\textbf{A} Input trace of a single synapse and voltage trace of a neuron. The emulator matches \loihi in both cases perfectly. Note that \loihi uses the voltage register to count refractory time, which results in a functionally irrelevant difference after a spike, e.g time step 17 in A2 (see Appendix \ref{subsubsec:voltage_memory}). \textbf{B} Network simulation with $400$ excitatory (indices $100-500$) and $100$ inhibitory (indices $0-100$) neurons. The network is driven by noise from an input population of $40$ Poisson spike generators with a connection probability of $0.05$. All spikes match exactly between the emulator and \loihi for all time steps. The figure shows the last 400 time steps from a simulation with $100\,000$ time steps.}
    \label{fig:simulations-match}
\end{figure*}

We now have all of the pieces required to understand and emulate a spiking unit from \loihi.
Evolving the differential equations for the current-based LIF model with the forward Euler method and using the appropriate rounding (see Section~\ref{sec:loihi:round_away}) and update schedule (see Section~\ref{sec:emulator:package} and Appendix \ref{sec:appendix:update-schedule}) is enough to exactly reproduce \loihi's behavior.
This procedure is summarized in Algorithm~\ref{algorithm:loihi} and an exact match between \loihi and an implementation for a single unit in \brian is shown in Figure~\ref{fig:simulations-match}A.
Please note that during the refractory period \loihi uses the voltage trace to count elapsed time (see Figure \ref{fig:simulations-match}A2, Appendix \ref{subsubsec:voltage_memory}), while in the emulator the voltage is simply clamped to zero.

\begin{algorithm}
    \SetAlgoLined
    \KwResult{Simulate one Loihi unit with one input synapse for $t_{max}$ time steps and read out state variables~($I$, $v$) and spikes ($\sigma$).}
    
    \texttt{\# Define round away from zero}\\
    $\text{rnd}(\cdot) \coloneqq \sign(\cdot) \ceil{ \abs{ \cdot }}$
    
    \texttt{\# Define input spike train}\\
    $S_t = \{0, 1\} \; \forall \; t \in \mathbb{N} \; \lvert \; t \leq t_{max}$
    
    \texttt{\# Define synaptic weight}\\
    $J := 2^{6+\Theta} \cdot \tilde w, \, \Theta \in [-8,7], \, \tilde w \in [-256, 255]$
    
    \texttt{\# Define threshold}\\
    $v_{th} := v_{mant} \cdot 2^6 ,\; v_{mant} \in [0, 131071]$
    
    \texttt{\# Define voltage and current decay}\\
    $\tau_v = 2^{12}/\delta^v,\; \delta^v \in [0, 4096]$ \\
    $\tau_I = 2^{12}/\delta^I,\; \delta^I \in [0, 4096]$
    
    \texttt{\# Initialise variables}\\
    $I_t, v_t, \sigma_t = 0 \; \forall \; t \in \mathbb{N} \; \lvert \; t \leq t_{max}$\\
    
    \texttt{\# Loop over simulation steps}\\
    \For{$t$ \texttt{from} $1$ \texttt{to} $t_{max}$}{
        
        \texttt{\# Spike input}\\
        $s \leftarrow S_t$
        
        \texttt{\# Update and read synaptic input}\\
        $I_{t} \leftarrow I_{t-1} - \text{rnd}(\frac{1}{\tau_I} I_{t-1}) + J \cdot s $
        
        \texttt{\# Update and read voltage}\\
        $v_{t} \leftarrow v_{t-1} - \text{rnd}(\frac{1}{\tau_v} v_{t-1}) + I_{t}$
        
        \texttt{\# Check threshold}\\
        \If{$v > v_{th}$}{
            \texttt{\# Read spike}\\
            $\sigma_t \leftarrow 1$
            
            \texttt{\# Reset voltage}\\
            $v_t \leftarrow 0$
        }
    }
    \caption{Loihi single neuron emulator}
    \label{algorithm:loihi}
\end{algorithm}

\newpage
\section{Network and plasticity}

We now have a working implementation of \loihi's spiking unit.
In the next step, we need to connect these units up into networks.
And if the network should be able to learn online, connections between units should be plastic.
In this section we review how weights are defined on \loihi and how learning rules are applied.
This includes the calculation of pre- and post-synaptic traces.
Based on this, we outline how these features are implemented in the emulator.

\subsection{Synaptic weights}
\label{subsec:synaptic_weights}

The synaptic weight consists of two parts, a weight mantissa $\tilde{w}$ and a weight exponent $\Theta$ and is of the form $\tilde{w} \cdot 2^{6 + \Theta}$.
However, in practice the calculation of the synaptic weight depends on bit shifts and its precision depends on a few parameters (see below).
The weight exponent is a value between $-8$ and $7$ that scales the weight mantissa exponentially.
Depending on the sign mode of the weight (excitatory, inhibitory, or mixed), the mantissa is an integer in the range $\tilde{w} \in [0,255]$, $\tilde{w} \in [-255,0]$, or $\tilde{w} \in [-256,254]$, respectively.
The possible values of the mantissa depend on the number of bits available for storing the weight and whether the sign mode is \textit{mixed} or not.
In particular, precision is defined as $2^{n_s}$, with
\begin{equation}\label{eq:num_weight_bits}
    n_{s} = 8 - (n_{wb} - \sigma_{\text{mixed}}).
\end{equation}
This can intuitively be understood with a few examples. 
If the weight bits for the weight mantissa are set to the default value of $n_{wb} = 8$ bits, it can store $256$ values between $0$ and $255$, i.e. the precision is then $2^{8-(8-0)}=2^0 = 1$.
If $n_{wb} = 6$ bits is chosen, we instead have a precision of $2^{8-(6-0)}=2^2 = 4$ meaning there are $64$ possible values for the weight mantissa, $\tilde{w} \in \{0, 4, 8, 16, ..., 252\}$.
If the sign mode is \emph{mixed}, i.e. $\sigma_{mixed}=1$, one bit is used to store the sign, which reduces the precision.
Mixed mode enables both positive and negative weights, with weight mantissa between $-256$ and $254$.
Assuming $n_{wb}=8$ in mixed mode, precision is $2^{8-(8-1)} = 2^1 = 2$ and $\tilde{w} \in \{-256, -254, ..., -4, -2, 0, 2, 4, ..., 254\}$.

\subsubsection{Weight initialisation}

While the user can define an arbitrary weight mantissa within the allowed range, during initialisation the value is rounded, given the precision, to the next possible value towards zero. 
This is achieved via bit shifting, that is the weight mantissa is shifted by
\begin{equation}
    \tilde{w}^{\text{shifted}} = (\tilde{w} \gg n_{s}) \ll n_{s},
\end{equation}
where $\gg$ and $\ll$ are a right and left shift respectively.
Afterwards the weight exponent is used to scale the weight according to
\begin{equation}
    J^{\text{scaled}} = \tilde{w}^{\text{shifted}} \cdot 2^{6 + \Theta}.
\end{equation}
This value cannot be greater than $21$ bits and is clipped if it exceeds this limit.
Note that this only happens in one case for $\tilde{w} = -256$ and $\Theta = 7$.
Finally the scaled value $J^{\text{scaled}}$ is shifted again according to
\begin{equation}
    J = (J^{\text{scaled}} \gg  6) \ll 6,
\end{equation}
where $J$ is the final weight. 

We provide a table with all $4096$ possible weights depending on the mantissa and the exponent in a Jupyter notebook\footnote{anonymized}.
These values are provided for all three sign modes.

\subsubsection{Plastic synapses}
\label{sec:weights-plastic}

In the case of a \textit{static} synapse, the initialised weight remains the same as long as the chip/emulator is running. 
Thus \textit{static} synapses are fully described by the details above.
For \textit{plastic} synapses, the weight can change over time.
This requires a method to ensure that changes to the weight adhere to its precision.

For \textit{plastic} synapses, \emph{stochastic rounding} is applied to the mantissa during each weight update.
Whether the weight mantissa is rounded up or down depends on its proximity to the nearest possible values above and below, i.e.
\begin{align} \label{eq:stochastic_rounding}
    \text{RS}_{2^{n_s}}(x) =
    \begin{cases}
      \sign(x) \cdot \lfloor \abs{x} \rfloor_{2^{n_s}} \qquad &\text{with probability} \; (2^{n_s} - (\abs{x} - \lfloor \abs{x} \rfloor_{2^{n_s}} )) / 2^{n_s}  \\
       \sign(x) \cdot (\lfloor \abs{x} \rfloor_{2^{n_s}} + 2^{n_s}) \qquad &\text{with probability} \; (\abs{x} - \lfloor \abs{x} \rfloor_{2^{n_s}} )) / 2^{n_s}
    \end{cases}
\end{align}
where $\lfloor \cdot \rfloor_{2^{n_s}}$ denotes rounding down to the nearest multiple of $2^{n_s}$.
After the mantissa is rounded, it is scaled by the weight exponent and the right/left bit shifting is applied to the result to compute the actual weight $J$.
How this is realised in the emulator is shown in Code~Listing~\ref{code:weights}.

To test that our implementation of the weight update for \textit{plastic} synapses matches \loihi for each possible number of weight bits, we compared the progression of the weights over time for a simple learning rule.
The analysis is described in detail in Appendix~\ref{sec:appendix:plastic-weight-update}.

\subsection{Pre- and post-synaptic traces}
\label{sec:learning-traces}

Pre- and post-synaptic traces are used for defining learning rules. \loihi provides two pre-synaptic traces $x_1$, $x_2$ and three post-synaptic traces $y_1$, $y_2$, $y_3$.
Pre-synaptic traces are increased by a constant value $\hat{x}_i$, for $i \in \{1, 2\}$, if the pre-synaptic neuron spikes.
The post-synaptic traces are increased by $\hat{y}_j$ for $j \in \{1, 2, 3\}$, accordingly.
So-called \emph{dependency factors} are available, indicating events like $x_0 = 1$ if the pre-synaptic neuron spikes or $y_0 = 1$ if the post-synaptic neuron spikes.
These factors can be combined with the trace variables by addition, subtraction, or multiplication.

A simple spike-time dependent plasticity (STDP) rule with an asymmetric learning window would, for example, look like $dw = x_1 \cdot y_0 - y_1 \cdot x_0$.
This rule leads to a positive change in the weight ($dw > 0$) if the pre-synaptic neuron fires shortly before the post-synaptic neuron (i.e. positive trace $x_1 > 0$ when $y_0 = 1$) and to a negative change ($dw < 0$) if the post-synaptic neuron fires shortly before the pre-synaptic neuron (i.e. positive trace $y_1 > 0$ when $x_0 = 1$). 
Thus, the time window in which changes may occur depends on the shape of the traces (i.e. impulse strength $\hat{x}_i$, $\hat{y}_i$; and decay $\tau_{x_i}$, $\tau_{y_j}$, see below).

For a sequence of spikes $s[t] \in \{0,1\}$, a trace is defined as
\begin{equation}\label{equation:synaptic_trace}
    x_i[t] = \alpha \cdot x_i[t-1]  + \hat{x}_i \cdot s[t],
\end{equation}
where $\alpha$ is a decay factor (\citealp[see][]{davies2018loihi}). 
This equation holds for presynaptic ($x_i$) and postsynaptic ($y_i$) traces.
However, in practice, on \loihi one does not set $\alpha$ directly but instead decay time constants $\tau_{x_i}$ and $\tau_{y_j}$. 

\begin{figure*}[ht]
    \centering
    \includegraphics[width=0.99\textwidth]{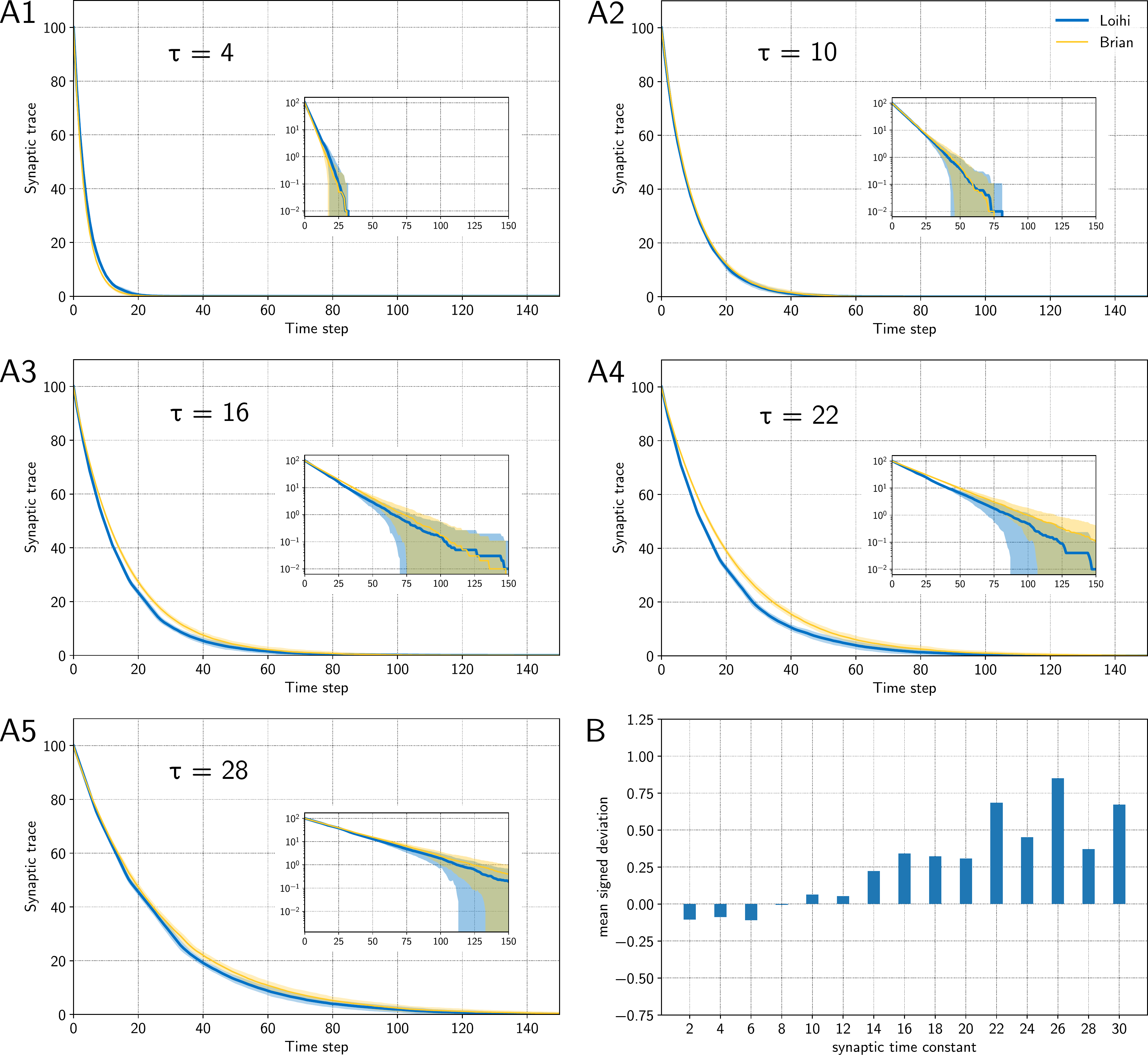}
    \caption{Deviations of the synaptic traces between \loihi and the emulator. \textbf{A} Synaptic traces for different synaptic time constants $\tau$. Averaged over $100$ trials each. The inlay shows the traces in a logarithmic scale. Blue indicates the trace from \loihi, yellow the trace from the emulator. \textbf{B} Mean signed deviation for different synaptic time constants $tau$ over $100$ trials each. For low $tau$ values, the emulator is slightly below the \loihi reference, whereas it lies slightly above the \loihi traces for higher values.}
    \label{fig:appendix:decay-deviation}
\end{figure*}

In the implementation of the emulator we again assume a first order approximation for synaptic traces, akin to synaptic input and voltage.
Under this assumption for the exponential decay, in Equation~\ref{equation:synaptic_trace} we replace $\alpha$ by
\begin{equation}
    \alpha(\tau_{x_i}) = 1 - \frac{1}{\tau_{x_i}}.
\end{equation}
Using this approximation gives reasonable results across a number of different $\tau_{x_i}$ and $\tau_{y_i}$ values (see Figure~\ref{fig:appendix:decay-deviation}).
While this essentially suffices, it could be improved by introducing an additional parameter, e.g. $\beta$, and optimising $\alpha(\tau_{x_i}, \beta)$.  

Note that we have integer precision again.
But different from the \textit{round away from zero} applied in the neuron model, here \textit{stochastic rounding} is used.
Since traces are positive values between $0$ and $127$ with precision $1$, the definition above in Equation~\ref{eq:stochastic_rounding} simplifies to the following
\begin{align}
    \text{RS}_{1, \ge 0}(x) =
    \begin{cases}
      \floor x \qquad &\text{with probability} \; 1 - (x - \floor x)  \\
      \floor x + 1 \qquad &\text{with probability} \; x - \floor x 
    \end{cases}
\end{align}
Since this rounding procedure is probabilistic and the details of the random number generator are unknown, rounding introduces discrepancies when emulating \loihi on the computer.
Further improvements are possible if more details of the chip's rounding mechanism were to be considered.

\subsection{Summary}

At this point we are able to connect neurons with synapses and build networks of neurons (see Figure~\ref{fig:simulations-match}B).
It was shown how the weights are handled, depending on the user defined number of weight bits or the sign mode.
In addition, using the dynamics of the pre- and post synaptic traces, we can now define learning rules.
Note that different from the neuron model, the synaptic traces cannot be reproduced exactly since the details of the random number generator, used for stochastic rounding, are unknown.
However, Figure~\ref{fig:emulator-learning} shows that the synaptic traces emulated in \brian are very close to the original ones in \loihi and that the behavior of a standard asymmetric STDP rule can be reproduced with the emulator.


\section{Loihi emulator based on Brian}

Here we provide an overview over the emulator package and show some examples and results.
This enables straightforward emulation of the basic features from \loihi as a sandbox for experimenters.
Note that we have explicitly not included routing and mapping restrictions, like limitations for the number of neurons or the amount of synapses, as these depend on constraints such as the number of used \loihi chips.

\subsection{The package}\label{sec:emulator:package}

The emulator package is available on \emph{PyPI}\footnote{\url{https://pypi.org/project/brian2-loihi/}} and can be installed using the \texttt{pip} package manager.
The emulator does not provide all functionality of the \loihi chip and software, but the main important aspects.
An overview over all provided features is given in Table~\ref{tab:features} in the appendix.
It contains six classes that extend the corresponding \brian classes.
The classes are briefly introduced in the following.
Further details can be taken from the code\footnote{anonymized}.

\subsubsection*{Network}

The \texttt{LoihiNetwork} class extends the \brian \texttt{Network} class.
It provides the same attributes as the original \brian class.
The main difference is that it initializes the default clock, the integration methods and updates the schedule when a \texttt{Network} instance is created.
Note that it is necessary to make explicitly use of the \texttt{LoihiNetwork}.
It is not possible to use \brian's \emph{magic network}.

Voltage and synaptic input are evolved with the forward Euler integration method, which was introduced in Section~\ref{seq:loihi:comp:neuro}.
Additionally a state updater was defined for the pre- and post-synaptic traces.

The default network update schedule for the computational order of the variables from \brian do not match the order of the computation on \loihi.
The \brian update schedule is therefore altered when initialising the \texttt{LoihiNetwork}, more details are given in Appendix \ref{sec:appendix:update-schedule}.

\subsubsection*{Neuron group}

The \texttt{LoihiNeuronGroup} extends \brian's \texttt{NeuronGroup} class.
Parameters of the \texttt{LoihiNeuronGroup} class are mostly different from the \brian class and are related to \loihi.
When an instance is created, the given parameters are first checked to match requirements from \loihi.
Finally, the differential equations to describe the neural system are shown in Code~Listing~\ref{code:neuron}.
Since \brian does not provide a \textit{round away from zero} functionality, we need to define it manually as an equation.

\subsubsection*{Synapses}
\begin{lstlisting}[
    style=mypython,
    frame=single,
    captionpos=b,
    label=code:neuron,
    float=t,
    numbers=left,
    caption={Neuron model equations of the voltage and the synaptic input for \brian. It contains a \emph{round away from zero} rounding.}
]
lif_equations = '''
    rnd_v = sign(v)*ceil(abs(v*1_tau_v)) : 1
    rnd_I = sign(I)*ceil(abs(I*1_tau_I)) : 1
    dv/dt = -rnd_v/ms + I/ms: 1 (unless refractory)
    dI/dt = -rnd_I/ms : 1
'''
\end{lstlisting}
The \texttt{LoihiSynapses} class extends the \texttt{Synapses} class from \brian.
Again, most of the \brian parameters are not supported and instead \loihi parameters are available.
When instantiating a \texttt{LoihiSynapses} object, the needed pre- and post-synaptic traces are included as equations (shown in Code~Listing~\ref{code:decay}) as theoretically introduced in Section~\ref{sec:learning-traces}.
Moreover, it is verified that the defined learning rule matches the available variables and operations supported by \loihi.
The equations for the weight update is shown in Code~Listing~\ref{code:weights}.

Since we have no access to the underlying mechanism and we cannot reproduce the pseudo-stochastic mechanisms exactly, we have to find a stochastic rounding that matches \loihi in distribution.
Note that on \loihi the same network configuration leads to reproducible results (i.e. same rounding).
Thus to compare the behavior of \loihi and the emulator, we simulate over a number of network settings and compare the distribution of the traces.
Figure~\ref{fig:emulator-learning}B shows the match between the distributions.
Note that with this, our implementation is always slightly different from the \loihi simulation, due to slight differences in rounding.
In Figure~\ref{fig:emulator-learning}C we show that these variations are constant and not diverging.
In addition, Figure~\ref{fig:emulator-learning}D shows that the principle behavior of a learning rule is preserved.
\begin{lstlisting}[
    style=mypython,
    frame=single,
    captionpos=b,
    label=code:decay,
    float=t,
    numbers=left,
    caption={Synaptic decay equation for \brian. Only the decay for $x1$ is shown, the decay for $x2$, $y1$, $y2$, $y3$ is applied analogously. It contains an approximation of the exponential decay and stochastic rounding.}
]
x1decay_equations = '''
    x1_new = x1 * (1 - (1.0/tau_x1)) : 1
    x1_int = int(x1_new) : 1
    x1_frac = x1_new - x1_int : 1
    x1_add_or_not = int(x1_frac > rand()) : 1 (constant over dt)
    x1_rnd = x1_int + x1_add_or_not : 1
    dx1/dt = x1_rnd / ms : 1 (clock-driven)
'''
\end{lstlisting}
\begin{lstlisting}[
    style=mypython,
    frame=single,
    captionpos=b,
    label=code:weights,
    float=t,
    numbers=left,
    caption={Weight equations for \brian. The first part creates variables that allow terms of the plasticity rule to be evaluated only at the $2^k$ time step. $dw$ contains the user defined learning rule. The updated weight mantissa is adapted depending on the number of weight bits, which determines the precision. The weight mantissa is rounded with \emph{stochastic rounding}. After clipping, the weight mantissa is updated and the actual weight is calculated.}
]
weight_equations = '''
    u0 = 1 : 1
    u1 = int(t/ms % 2**1 == 0) : 1
    ...
    u9 = int(t/ms % 2**9 == 0) : 1
    
    dw_rounded = int(sign(dw)*ceil(abs(dw))) : 1
    quotient = int(dw_rounded / precision) : 1
    remainder = abs(dw_rounded) % precision : 1
    prob = remainder / precision : 1
    add_or_not = sign(dw_rounded) * int(prob > rand()) : 1 (constant over dt)
    dw_rounded_to_precision = (quotient + add_or_not) * precision : 1
    w_updated = w + dw_rounded_to_precision : 1
    w_clipped = clip(w_updated, w_low, w_high) : 1
    dw/dt = w_clipped / ms : 1 (clock-driven)
    
    w_act_scaled = w_clipped * 2**(6 + w_exp) : 1
    w_act_scaled_shifted = int(floor(w_act_scaled / 2**6)) * 2**6 : 1
    w_act_clipped = clip(w_act_scaled_shifted, -limit, limit) : 1
    dw_act/dt = w_act_clipped / ms : 1 (clock-driven)
    
    dx0/dt = 0 / ms : 1 (clock-driven)
    dy0/dt = 0 / ms : 1 (clock-driven)
'''
\end{lstlisting}
\subsubsection*{State monitor \& Spike monitor}

The \texttt{LoihiStateMonitor} class extends the \texttt{StateMonitor} class from \brian, while the \texttt{LoihiSpikeMonitor} class extends the \texttt{SpikeMonitor} class.
Both classes support the most important parameters from their subclasses and update the schedule for the timing of the probes.
This schedule update avoids shifts in the monitored variables, compared to \loihi.

\subsubsection*{Spike generator group}

The \texttt{LoihiSpikeGeneratorGroup} extends the \texttt{SpikeGeneratorGroup} class from \brian.
This class only reduces the available parameters to avoid that users unintentionally change variables which would cause an unwanted emulation behavior.

\subsection{Results}

\begin{figure*}[t]
    \centering
    \includegraphics[width=0.99\textwidth]{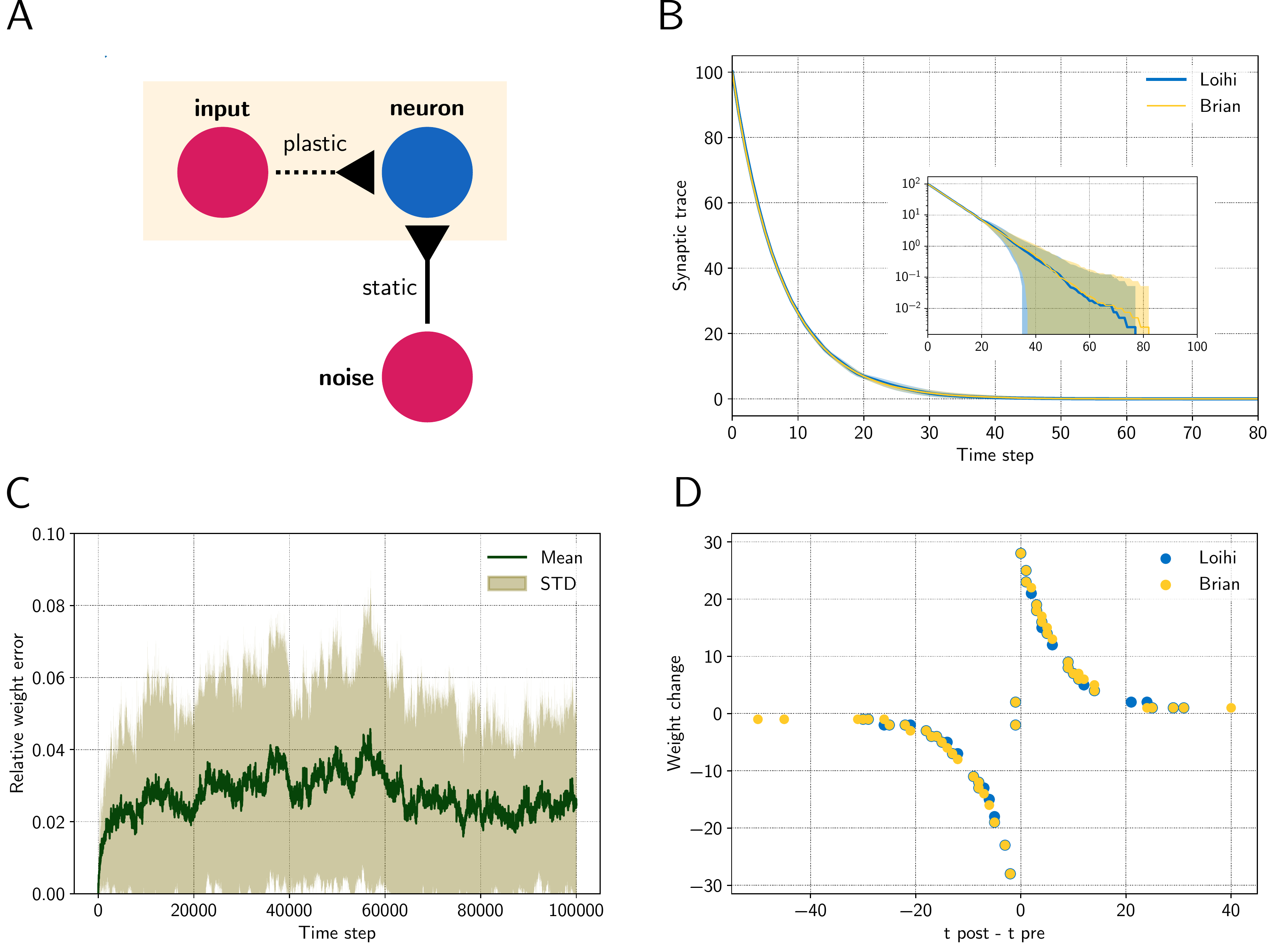}
    \caption{Comparing a STDP learning rule performed with the emulator and with \loihi. \textbf{A} Sketch showing the setup. \textbf{B} Synaptic trace for many trials showing the arithmetic mean and standard deviation. The inset shows the same data in a logarithmic scale. Note that every data point smaller than $10^{0}$ shows the probability of rounding values between $0$ and $1$ up or down. \textbf{C} Relative difference $| \tilde w_L - \tilde w_B | / \tilde w_{\text{max}}$ for the plastic weight between the emulator, $\tilde w_B$, and the \loihi implementation, $\tilde w_L$, for $50$ simulations, $\tilde w_{\text{max}} = 255$. \textbf{D} STDP weight change in respect to pre- and post-synaptic spike times, data shown for time steps $0-2000$ for visualisation purposes.}
    \label{fig:emulator-learning}
\end{figure*}

To demonstrate that the \loihi emulator works as expected, we provide three examples covering a single neuron, a recurrently connected spiking neural network, and the application of a learning rule.
All three examples are available as \texttt{jupyter} notebooks\footnote{anonymized}.

\subsubsection*{Neuron model}

In a first test, we simulated a single neuron.
The neuron receives randomly timed excitatory and inhibitory input spikes.
Figure~\ref{fig:simulations-match}A1 shows the synaptic responses induced by the input spikes for the simulation using the \loihi chip and the \brian emulator.
The corresponding voltage traces are shown in Figure~\ref{fig:simulations-match}A2.
As expected, the synaptic input as well as the voltage match perfectly between both hardware types.

\subsubsection*{Network}

In a second approach we applied a recurrently connected network of $400$ excitatory and $100$ inhibitory neurons with log-normal weights.
The network gets noisy background input from $40$ Poisson generators that are connected to the network with a probability of $0.05$.
As already shown by others, this setup leads to a highly chaotic behavior \citep{london2010sensitivity, brunel2000dynamics, van1996chaos, sompolinsky1988chaos}.
Despite the chaotic dynamics, spikes, voltages and synaptic inputs match perfectly for all neurons and over the whole time.
The spiking pattern of the network is shown in Figure~\ref{fig:simulations-match}B.
All yellow (\brian) and blue (\loihi) dots match perfectly.

\subsubsection*{Learning}

In the last experiment, we applied a simple STDP learning rule, as introduced in Equation~\ref{eq:stdp}, at a single plastic synapse.
The experiment is sketched in Figure~\ref{fig:emulator-learning}A.
One spike generator, denoted \emph{input}, has a plastic connection to a neuron with a very low weight ($\tilde w = 128$, $\Theta = -6$), such that it has a negligible effect on the post-synaptic neuron.
Another spike generator, denoted \emph{noise}, has a large but static weight ($\tilde w = 254$, $\Theta = 0$) to reliably induce post-synaptic spikes.
Figure~\ref{fig:emulator-learning}B compares the distribution of traces between the emulator and \loihi.
For this $400$ trials were simulated.

We chose an asymmetric learning window for the STDP rule.
The learning rule uses one pre-synaptic trace $x_1$ ($\hat x_1 = 120$, $\tau_{x_1} = 8$) and one post-synaptic trace $y_1$ ($\hat y_1 = 120$, $\tau_{y_1} = 8$).
In addition the dependency factors $x_0 \in {0,1}$ and $y_0 \in {0,1}$ are used, which indicate a pre- and post-synaptic spike respectively.
Using these components, the learning rule is defined as
\begin{equation}\label{eq:stdp}
    dw = 2^{-2} \cdot x_1 \cdot y_0 - 2^{-2} \cdot x_0 \cdot y_1.
\end{equation}

Due to the stochastic rounding of the traces, differences in the weight changes occur, which are shown in Figure~\ref{fig:emulator-learning}C.
Fortunately, the differences in the weight changes remain on a constant level and do not diverge, even over long simulation times, e.g. $100\,000$ steps.
Despite these variations, the STDP learning window of the emulator reproduces the behavior of the \loihi learning window, as shown in Figure~\ref{fig:emulator-learning}D.

\section{Discussion}

This study was motivated by two goals.
We hope to simplify the transfer of models to \loihi and therefore developed a \loihi emulator for \brian, featuring many functionalities of the \loihi chip.
In the process of developing the emulator, we aimed to provide a deeper understanding of the functionality of the neuromorphic research chip \loihi by analysing its neuron and synapse model, as well as synaptic plasticity.

We hope that the analysis of \loihi's spiking units has provided some insight into how \loihi computes.
With the numerical integration method, numerical precision and related rounding method, as well as the update schedule, we were able to walk from the LIF neuron model down to the computations performed. 
For neurons and networks without plasticity we are able to emulate \loihi without error.
Analysing and implementing synaptic plasticity showed that, due to stochastic rounding, it is not possible to exactly replicate trial by trial behavior when it comes to learning.
However, on average the weight changes induced by a learning rule are preserved.

The main benefit of the \texttt{Brian2Loihi} emulator lies in lowering the hurdle for the experimenter.
Especially in neuroscience, many scientists are accustomed to neuron simulators and in particular \brian is widely used.
The emulator can be used for simple and fast prototyping, making a deep dive into new software frameworks and hardware systems unnecessary.
In addition, hardware specific complications, like distributing neurons to cores, or constraints like potential limits on the number of available neurons or synapses, or on the speed or size of read-out, do not occur in the emulator.
While this will surely improve with new generations of hardware and software in the upcoming years, they can already be ignored by using the emulator.

At this point it is important to note that not all \loihi features are included in the emulator, yet.
In particular, the homeostasis mechanism, rewards, and tags for the learning rule are not included.
In Table~\ref{tab:features} we provide a comparison of all functionalities from \loihi with those available in the current state of the emulator.
Development of this emulator is an open source project and we expect improvements and additions with time.

An important vision for the future is to flexibly connect front-end development environments (e.g. \brian, NEST, Keras, TensorFlow) with various back-ends, like neuromorphic platforms (e.g. \loihi, SpiNNaker, BrainScaleS, Dynap-SE) or emulators for these platforms.
PyNN \citep{davison2009pynn} is such an approach to unify different front-ends and back-ends in a more general way.
Nengo \citep{Bekolay2014}, as another approach, does not provide the use of other simulators, but allows several back-ends and focuses on higher level applications \citep{dewolf2020nengo}.
NxTF \citep{rueckauer2021nxtf} is an API and compiler aimed at simplifying the efficient deployment of deep convolutional spiking neural networks on \loihi using an interface derived from Keras.
We think that ideally, one could continue to work in their preferred front-end environment while a package maps their code to existing chips or computer-based emulators of these chips.
We expect an interface along these lines will play an important role in the future of neuromorphic computing and want to contribute to this development with our \texttt{Brian2Loihi} emulator.

At least for now, with an emulator at hand, it is easier to prototype network models and assess whether an implementation on \loihi is worth considering.
When getting started with neuromorphic hardware, to e.g. scale up models or speed up simulations, researchers familiar with \brian can directly deploy models prepared with the emulator.
We hope that with this, others may find a smooth entry into the quickly emerging field of neuromorphic computing.

\section{Acknowledgements}

The work received funds by the Intel Corporation via a gift without restrictions. ABL currently holds a Natural Sciences and Engineering Research Council of Canada PGSD-3 scholarship. We would like to thank
Jonas Neuhöfer, Sebastian Schmitt, Andreas Wild, and Terrence C. Stewart for valuable discussions and input.

\newpage


\section{Appendix}


\subsection{Loihi neuron model}\label{sec:appendix:lif-variant}

Computational units on \loihi communicate via spikes.
They can be connected up to form networks, each unit both sending and receiving spikes from some subset of the other units.
Like neurons in the brain, a unit emits a spike if its internal variable reaches a certain threshold.
The spike is then transmitted to all units with a direct incoming connection from the one that spiked. 
This induces a change in the receiving units' internal variable.
At every time step, the internal variable of all units' decays towards zero, counteracting any input received.
And after spiking, the internal variable is reset to zero.
In terms of the brain, each computational unit on \loihi implements a simple model of a spiking neuron, in particular a variant of the leaky integrate and fire neuron model, which is based on a simple resistor-capacitor (RC) circuit.
Readers are encouraged to consult the first chapter of \citet{gerstner2014neuronal} for a more detailed treatment.

\subsubsection{Voltage}
We refer to the standard leaky integrate and fire neuron, as it is defined in \citet{gerstner2014neuronal}.
In this model, the difference in electric potential between the interior and the exterior of a neuron, the so-called membrane potential, evolves according to

\begin{equation}\label{eq:appendix:voltage:brian}
    {\tau_v} \frac{dv}{dt} = - [v(t) - v_{rest}] + R I(t),
\end{equation}

where $v$ is the voltage across the membrane, $\tau_v$ is the membrane time constant of the neuron, $v_{rest}$ is the resting potential, $I$ is the input current and $R$ is the resistance of the membrane. 
Whenever the membrane potential reaches threshold $v_i^{th}$ it is reset to $v_{rest}$.

\citet{davies2018loihi} present the following variant of the standard LIF model which forms the basis for \loihi's computational units

\begin{equation}\label{eq:appendix:loihi:voltage:basis}
    \frac{dv_i}{dt} = - \frac{1}{\tau_v} v_i(t)  + I_i(t) - v_i^{th} \sigma_i(t),
\end{equation}

where $v$ is the voltage across the membrane, $\tau_v$ is the time constant for voltage decay, $I$ is in this case an input variable, $v_i^{th}$ is the threshold voltage to spike, and $\sigma(t)$ indicates whether the neuron fired a spike at time~$t$.

There are a few differences that are worth noting.
In the \loihi variant, the resting potential is zero.
The membrane time constant $\tau_v$ applies only to the voltage decay and not to the input variable $I$.
In effect, the resistance and the time constant are implicit in the input variable I as connection weight.

Further, resetting after a spike is included directly in the differential equation.
The final term subtracts the threshold voltage $v_i^{th}$ at the time of a spike.
This is a matter of notation and can be written as such, or with a separate reset condition $v_i \to 0$ applied at the time of each spike, as in \citet[][]{gerstner2014neuronal}.

\subsubsection{Derivation of synaptic response}
\label{subsec:appendix:synaptic_response}

To keep this paper self-contained, here we derive the synaptic response of a \loihi unit to an incoming spike train. 
For the reader's convenience, we repeat the definition from Equation \ref{eq:current:loihi:paper:original} in the main text here and then with a few steps obtain the result from Equation \ref{eq:loihi:input:result}.

\textbf{Definition.} The \textit{synaptic response} is given by 
\begin{equation}
    I_i(t) = \sum_j J_{ij} (\alpha_I * \sigma_j)(t) + I_i^\text{bias},
\end{equation}
where $J_{ij}$ is the weight from unit $j$ to $i$, $I_i^\text{bias}$ is a constant bias input, and the spike train $\sigma_j$ of unit $j$ is convolved with the \textit{synaptic filter impulse response} $\alpha_I$, given by 
\begin{equation}
    \alpha_I(t) = \exp\left(-\frac{t}{\tau_I}\right)\,H(t),
\end{equation}
where $\tau_I$ is the time constant of the synaptic response and $H(t)$ the unit step function.
Note we define $\alpha_I(t)$ differently here than in \citet{davies2018loihi} (see Appendix~\ref{subsubsec:appendix:synaptic_filter} for details).

\textbf{Definition.} The \textit{unit step function} $H : \mathbbm{R} \to \mathbbm{R}$ is given by 
\begin{equation}\label{eq:heaviside}
H(x) = 
    \begin{cases}
        1, & x \ge 0\\
        0, & x < 0.\\
    \end{cases}  
\end{equation}

\textbf{Definition.} The \textit{Dirac delta} is a tempered distribution $\delta \in \mathcal {S}'(\mathbbm{R})$, with  $\delta : \mathcal {S}(\mathbbm{R}) \to \mathbbm{C}, \; \varphi \mapsto \langle \delta, \varphi \rangle$ where
\begin{equation}\label{eq:dirac}
    \langle \delta, \varphi \rangle := \int_{-\infty}^{\infty} \delta(x) \varphi(x) \, dx := \varphi(0)
\end{equation}
for all Schwartz functions $\varphi \in \mathcal {S}(\mathbbm{R})$. Here we extend the definition such that $\delta : f \to f(0)$ for arbitrary, everywhere-defined $f : \mathbbm{R} \to \mathbbm{R}$.

\textbf{Definition.} We define the translation of $\delta$ by $a$, denoted $\delta_a$, as the distribution $\tau_a \delta : \mathcal{S}(\mathbbm{R}) \to \mathbbm{C}$ with
\begin{equation}
    \tau_a \delta (\varphi) := \langle \delta_a, \varphi \rangle = \int_{-\infty}^{\infty} \delta(x - a) \varphi(x) \, dx 
\end{equation}
and again extend this notion to arbitrary, everywhere-defined $f : \mathbbm{R} \to \mathbbm{R}$.

\textbf{Lemma 1.} (\textit{translation property}) $\tau_a \delta (f) = f(a)$, for $a \in \mathbbm{R}$ and $f : \mathbbm{R} \to \mathbbm{R}$.

\textbf{Proof.} Let $f : \mathbbm{R} \to \mathbbm{R}$. Then 
\begin{equation}
    \tau_a \delta (f) = \int_{-\infty}^{\infty} \delta(x - a) f(x) \, dx = \int_{-\infty}^{\infty} \delta(x) f(x + a) \, dx = f(0 + a) = f(a)
\end{equation}
\begin{flushright}
    $\blacksquare$
\end{flushright}

\hypertarget{corollary:dirac}{\textbf{Corollary 1.}} As a sum of Dirac deltas, $\sigma_i$ can be understood as the following linear functional 
\begin{equation}
\sigma_i := \sum_k \tau_{t_{i,k}} \delta \; :  \;  \varphi \mapsto \mathbbm{C},
\end{equation}
with
\begin{equation}
\langle \sigma_i, \varphi \rangle := \langle \sum_k \delta_{t_{i,k}}, \varphi \rangle = \sum_k \langle \delta_{t_{i,k}}, \varphi \rangle = \sum_k \varphi(t_{i,k}), \qquad \varphi \in \mathcal{S(\mathbbm{R})},
\end{equation}
and again we extend this notion from the space of tempered distributions to $\sigma_i$ for arbitrary, everywhere-defined $f$.

\textbf{Definition.} The \textit{convolution} between the Dirac delta distribution and a function is to be understood in the following sense
\begin{equation}
    (\delta * f)(x) := \langle \delta, \tau_{x} \tilde f \rangle = \int_{-\infty}^{\infty} \delta(y) f(x - y) \, dy = \int_{-\infty}^{\infty} \delta(x - y) f(y) \, dy
\end{equation}
where $\tilde f(x) = f(-x)$. 

\textbf{Lemma 2.} $(\delta * f)(x) = f(x)$.

\textbf{Proof.}
Using $\delta(x) = \delta(-x)$ (E) and the translation property of the Dirac delta function (T) from Lemma~1 we have
\begin{equation}
    (\delta * f)(x) := \int_{-\infty}^{\infty} \delta(x - y) f(y) \, dy \stackrel{\tiny E}{=}
    \int_{-\infty}^{\infty} \delta(y - x) f(y) \, dy = 
    \tau_x \delta (f) \stackrel{\tiny T}{=} f(x).
\end{equation}
\begin{flushright}
    $\blacksquare$
\end{flushright}

\textbf{Claim.} The synaptic input $I_i(t)$ for unit $i$ is given by
$$
I_i(t) = \sum_j J_{ij} \sum_k   \exp\left(\frac{t_{j,k} - t}{\tau_I}\right)H(t-t_{j,k}) + I_i^\text{bias}.
$$
\textbf{Proof.} Applying the definition of convolution (D), linearity of the integral operator (L), the translation property of the Dirac delta function (T), and using that $\delta(x) = \delta(-x)$ (E) we have 
\begin{alignat}{2}
    (\alpha_I * \sigma_j)(t) \, &\stackrel{\tiny D}{=}& \; & \int_{-\infty}^{\infty} \alpha_I(s)  \, \sigma_j(t - s) \, ds \\
    &=& \; & \int_{-\infty}^{\infty} \alpha_I(s)  \sum_k \delta(t - t_{j,k} - s) \, ds \\
    &\stackrel{\tiny L}{=}& \; & \sum_k \int_{-\infty}^{\infty} \alpha_I(s)  \delta(t - t_{j,k} - s) \,  ds \\
    &\stackrel{\tiny E}{=}& \; & \sum_k \int_{-\infty}^{\infty} \alpha_I(s)  \delta(s - (t - t_{j,k})) \, ds \\
    &=& \; & \sum_k \tau_{t - t_{j,k}} \delta( \alpha_I) \\
    &\stackrel{\tiny T}{=}& \; &  \sum_k \alpha_I(t - t_{j,k})
\end{alignat}

\noindent With this, we can write the synaptic input (Equation \ref{eq:current:loihi:paper:original}) as
\begin{alignat}{2}
    I_i(t) &=& \; & \sum_j J_{ij} \, \sum_k \alpha_I(t - t_{i,k}) + I_i^\text{bias} \\
    &=& \; & \sum_j J_{ij} \sum_k  \exp\left(\frac{t_{j,k} - t}{\tau_I}\right)H(t-t_{j,k}) + I_i^\text{bias}. \label{eq:syn_input}
\end{alignat}
\begin{flushright}
    $\blacksquare$
\end{flushright}

We see the input can be written as a sum of exponentially decaying functions with amplitude $J_{ij}$ beginning at the time of each spike $t_{j,k}$. 

\subsubsection{Definition of the synaptic filter impulse response} \label{subsubsec:appendix:synaptic_filter}

\citet[][]{davies2018loihi} defined the \textit{synaptic filter impulse response} as
\begin{equation}
    \alpha_I^{orig}(t) = \frac{1}{\tau_I} \exp\left(-\frac{t}{\tau_I}\right)\,H(t).
\end{equation}
Note that we have omitted the factor of $1 / \tau_I$ in our definition, in particular we defined
\begin{equation}
    \alpha_I(t) = \exp\left(-\frac{t}{\tau_I}\right)\,H(t).
\end{equation}
We prefer this formulation as the results obtained match exactly with the \loihi documentation.
If, however, the factor of $1/\tau_I$ is included, the factor is carried through to Equation \ref{eq:loihi:input:euler:delta}. Namely it becomes
\begin{equation}
    I[t] = I[t-1] \cdot (2^{12} - \delta^I) \cdot 2^{-12} + \frac{J}{\tau_I} \cdot s[t]
\end{equation}

where we see there is an extra factor of $1/\tau_I$ multiplied by the weight $J$. 
The definition from \citet[][]{davies2018loihi} and the \nxsdk documentation can be reconciled by replacing this extra factor of $1/\tau_I$ with a static factor $2^{6}$ and then considering the weight to be $J = \tilde w \cdot 2^{\Theta}$ instead of $J = \tilde w \cdot 2^{6 + \Theta}$.

\subsection{Miscellaneous implementational details}

\subsubsection{Brian state update schedule}\label{sec:appendix:update-schedule}

In \brian the \texttt{network} class is the main class of a simulation.
All containing objects like neurons, synapses, monitors, poisson generators, are added to that \texttt{network} object.
Each of these objects have a \textit{when} attribute.
The \texttt{network} class decides in which order containing objects are updated depending on their \textit{when} attribute.
For this decision a schedule is defined, given as a string list.
The default schedule is \texttt{['start', 'groups', 'thresholds', 'synapses', 'resets', 'end']}.

We observed that \loihi implements a schedule where first the synapses are updated and afterwards the neuron groups.
In \brian the evaluation is performed in opposite order, which results in a shift between \loihi and the emulator.
We therefore changed the \brian schedule to \texttt{['start', 'synapses', 'groups', 'thresholds', 'resets', 'end']}, i.e. the synapse update is pulled in front of groups.

Additionally the time when the synaptic monitor is evaluated is different in \loihi.
For the emulator, we also needed to adjust these.
This is done by changing the monitors \textit{when} flag from the default \textit{start} to \textit{synapses} for the synaptic input and all pre- and post-synaptic trace variables.
For probing the voltage and weight the \textit{when} attribute was changed to \textit{end}.
The same holds for probing spikes with the spike monitor.
Moreover, the poisson generators \textit{when} flag has to be changed from the default \textit{thresholds} to \textit{start} to ensure Poisson spikes are given at the beginning of the current time step and are propagated through the simulation schedule.

\subsubsection{Voltage memory used to count refractory time} \label{subsubsec:voltage_memory}

Note that \loihi sets the voltage of a neuron to a non zero value if the neuron has spiked.
The memory for storing the voltage is used for counting while the neuron is in refractory state.
This causes a deviation between the emulator and \loihi for the voltage, which is only due to technical reasons and has no functional effect.

\newpage

\subsection{Plastic weight update with stochastic rounding}
\label{sec:appendix:plastic-weight-update}

\begin{figure*}[ht]
    \centering
    \includegraphics[width=0.99\textwidth]{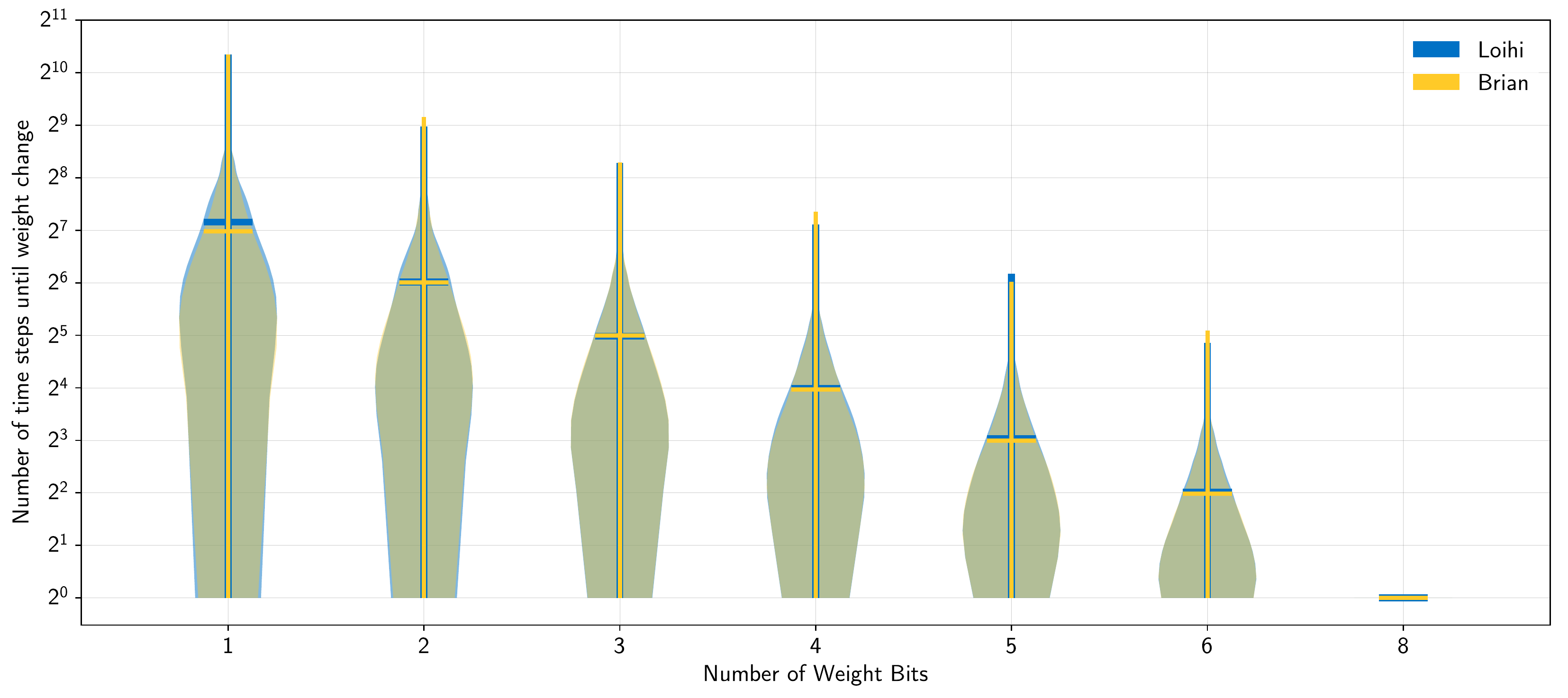}
    \caption{Distribution of the weight change for different number of weight bits. The weight mantissa is increased by~$1$ in every time step. Due to stochastic rounding, this change may then be rounded up or down. Shown is the distribution of the number of time steps until a weight change occurs. For each number of weight bits, $8000$ weight changes were sampled for \loihi and the emulator. The emulator implementation matches \loihi well.}
    \label{fig:appendix:times-weight-change-distribution}
\end{figure*}

If the weight is updated by a learning rule, the weight mantissa needs to be updated according to the given precision, as described in Section~\ref{sec:weights-plastic}.
The precision is determined by the available number of bits, which can be chosen by the user, and in addition depends on the sign mode.
To test the implementation in the emulator, we compared its behavior to \loihi for each possible number of weight bits.
In particular, for an excitatory plastic synapse we increased the weight mantissa by one at each time step (via learning rule $dw = u_0$) and measured the actual weight after the update (i.e. rounding and shifting).
Our expectation was that for \textit{stochastic rounding} to the nearest $2^{n_s}$, the average number of time steps required until a weight change takes place should be equal to $2^{n_s}$.
This is because the probability of rounding up from a given weight mantissa, e.g. $\tilde w := k \cdot 2^{n_s}$, when $1$ is added can be calculated from Equation~\ref{eq:stochastic_rounding} as
\begin{equation*}
    (\abs{w} - \lfloor \abs{w} \rfloor_{2^{n_s}} )) / 2^{n_s} = ((k \cdot 2^{n_s} + 1) + k \cdot 2^{n_s})/k \cdot 2^{n_s} = 1 / 2^{n_s}.
\end{equation*}
As expected, the results match \loihi's behavior nicely, as seen in Figure~\ref{fig:appendix:times-weight-change-distribution}, confirming the validity of our implementation.

\newpage

\subsection{Pre- and post-synaptic decay deviations}
\label{sec:appendix:synaptic-decay-deviations}

\newpage

\subsection{Emulator features}
\label{sec:appendix:emulator-features}

\begin{table}[!ht]
    \centering
    \begin{tabular}{p{7cm}|c}
        \textbf{Loihi} & \textbf{Emulator}  \\
        \hline
        neurons &  \\
        \hline
        current impulse/decay & \checkmark \\
        voltage impulse/decay & \checkmark \\
        bias input & (\checkmark) \\
        homeostasis (threshold adaption) & - \\
        random noise for current & - \\
        random noise for voltage & (\checkmark) \\ 
        multi-compartment neurons & (\checkmark) \\
        \hline
        connections &  \\
        \hline
        weight mantissa/exponent & \checkmark \\
        weight precision & \checkmark \\
        synaptic delay & \checkmark \\
        box-synapse & - \\
        \hline
        learning &  \\
        \hline
        presynaptic spike & \checkmark \\
        $1^{st}$ presynaptic trace & \checkmark \\
        $2^{nd}$ presynaptic trace & \checkmark \\
        postsynaptic spike & \checkmark \\
        $1^{st}$ postsynaptic trace & \checkmark \\
        $2^{nd}$ postsynaptic trace & \checkmark \\
        $3^{nd}$ postsynaptic trace & \checkmark \\
        synaptic weight as variable & \checkmark \\
        reward spike & - \\
        reward trace & - \\
        tag & - \\
        plastic synaptic delay & - \\
        learning epoch & (\checkmark) \\
        \hline
        probes &  \\
        \hline
        probe variables & \checkmark \\
        probing conditions & (\checkmark) \\
    \end{tabular}
    \caption{Features of \loihi compared with the emulator (version 0.5.2). Check marks in brackets are not fully supported or can manually be included using core \brian functionality.}
    \label{tab:features}
\end{table}

\newpage


\printbibliography

\end{document}